\documentclass[sigconf]{acmart}

\AtBeginDocument{%
  \providecommand\BibTeX{{%
    \normalfont B\kern-0.5em{\scshape i\kern-0.25em b}\kern-0.8em\TeX}}}

\usepackage{wrapfig}

\usepackage{caption}
\usepackage{subcaption}

\usepackage{amsmath}
\usepackage{mathtools}

\newcommand{\Lagr}{\mathcal{L}}
\usepackage{algorithm,algpseudocode}
\usepackage{microtype}

\usepackage{ragged2e}
\usepackage{multirow}
\usepackage{graphicx}

\usepackage{tikz}

\begin{document}
\captionsetup[table]{skip=0pt}
\captionsetup[figure]{skip=0pt}
\title{Examining the Role and Limits of Batchnorm Optimization to Mitigate Diverse Hardware-noise in In-memory Computing}


\author{Abhiroop Bhattacharjee}
\affiliation{%
  \institution{Yale University}
  \city{New Haven}
  \country{USA}
  }
\email{abhiroop.bhattacharjee@yale.edu}

\author{Abhishek Moitra}
\affiliation{%
  \institution{Yale University}
  \city{New Haven}
  \country{USA}
  }
\email{abhishek.moitra@yale.edu}

\author{Youngeun Kim}
\affiliation{%
  \institution{Yale University}
  \city{New Haven}
  \country{USA}
  }
\email{youngeun.kim@yale.edu}

\author{Yeshwanth Venkatesha}
\affiliation{%
  \institution{Yale University}
  \city{New Haven}
  \country{USA}
  }
\email{yeshwanth.venkatesha@yale.edu}

\author{Priyadarshini Panda}
\affiliation{%
  \institution{Yale University}
  \city{New Haven}
  \country{USA}
  }
\email{priya.panda@yale.edu}



\begin{abstract}

In-Memory Computing (IMC) platforms such as analog crossbars are gaining focus as they facilitate the acceleration of low-precision Deep Neural Networks (DNNs) with high area- \& compute-efficiencies. However, the intrinsic non-idealities in crossbars, which are often non-deterministic and non-linear, degrade the performance of the deployed DNNs. In addition to quantization errors, most frequently encountered non-idealities during inference include crossbar circuit-level parasitic resistances and device-level non-idealities such as stochastic read noise and temporal drift. In this work, our goal is to closely examine the distortions caused by these non-idealities on the dot-product operations in analog crossbars and explore the feasibility of a nearly \textit{training-less} solution via crossbar-aware fine-tuning of batchnorm parameters in real-time to mitigate the impact of the non-idealities. This enables reduction in hardware costs in terms of memory and training energy for IMC noise-aware retraining of the DNN weights on crossbars.

\end{abstract}





\maketitle

\section{Introduction}
\label{sec:intro}
\vspace{-1mm}
Deep Neural Networks (DNNs) have found ubiquitous applications in a broad spectrum of tasks ranging from computer vision, voice recognition to natural language processing \cite{alzubaidi2021review}. In the context of hardware implementation of DNNs for resource-constrained edge platforms, In-Memory Computing (IMC) systems, such as analog crossbar-arrays that alleviate the `memory-wall' bottleneck of von-Neumann architectures are gaining popularity \cite{sebastian2020memory}. Analog crossbars are based on various memristive devices such as- Resistive Random-access Memories (RRAMs), Phase Change Memories (PCMs), Ferroelectric Field-effect Transistors (FeFETs) that are widely researched for compact, energy-efficient and accurate inference of low-precision DNNs with high throughput \cite{chakraborty2020pathways}. 

Analog IMC platforms are susceptible to various non-idealities arising from the device-to-device variations and temporal conductance drift of the memristive devices as well as from the parasitic resistances of the metallic interconnects in the crossbar-arrays \cite{chakraborty2020geniex, jain2020rxnn, sun2019impact, antolini2023combined}. These non-idealities interfere with the weights of a DNN (programmed as conductances of the memristive devices) leading to inaccurate dot-product computations, thereby causing a significant loss in the classification accuracy of the mapped DNNs on crossbars. In recent years, several noise models \cite{rasch2023hardware, antolini2023combined, zhang2020mitigate} and crossbar-realistic DNN accuracy evaluation frameworks (Neurosim \cite{peng2020dnn+}, RxNN \cite{jain2020rxnn}, GenieX \cite{chakraborty2020geniex}, Memtorch \cite{lammie2020memtorch}) have been proposed that are integrated with hardware-aware re-training or fine-tuning of GPU-trained DNN weights, to mitigate performance losses during inference due to non-idealities as well as quantization of weights and activations. However, this hardware-integrated re-training can lead to a huge increase in the overall training cost in terms of GPU-hours \cite{roy2021txsim, negi2021nax, peng2020dnn+}. Moreover, owing to limited retention capabilities of memristive devices such as RRAMs and FeFETs due to temporal conductance drift, they need to be periodically re-programmed to prevent accuracy degradation during inference \cite{byun2022recent}. Note, memristive devices entail higher programming energy costs than standard SRAM cells. For works that support on-chip training and inference using realistic IMC crossbar platforms for forward and backward propagation stages, fine-tuning (or re-programming) the weights (or memristive conductances) of a pre-trained DNN on hardware to mitigate accuracy losses due to noise and quantization effects becomes highly compute- \& memory-intensive \cite{peng2020dnn+, jin2021rehy, rasch2021flexible}. This is due to the requirement to compute \& store gradients of weights and activations during the backpropagation stage. 

Considering the above challenges, in this work, we pose a realistic question- \textit{Given the underlying analog IMC architecture and its noise characteristics, is real-time crossbar-aware training of weights always necessary to preserve the accuracy of a convolutional DNN inferred on crossbars?} Here, we evaluate the scope of a nearly \textit{training-less} solution for the pre-trained weights of the convolutional layers to be reused across diverse hardware-noise existing in analog IMC inference workloads by adapting the batchnorm parameters. We carefully study and quantify the distortion of DNN weights on crossbars caused by various types of hardware noise, based on which we evaluate the impact of tuning batchnorm parameters (keeping the weights frozen in the convolution layers that are mapped onto the crossbars) in mitigating the non-linearities induced by the noise. Through this work, we wish to draw the attention of the research community towards the fact that it is not always worthwhile to explore costly hardware noise-aware weight fine-tuning/re-programming solutions for accurate inference of DNNs on non-ideal IMC crossbars. Instead, we show that simple noise-aware fine-tuning of batchnorm parameters can be a very effective and hardware-friendly solution.

In summary, the key contributions of this work are as follows:

\begin{itemize}
    \item This work deals with the implementation of the convolution layers of a pre-trained DNN model using non-ideal analog crossbars. We propose that hardware noise-aware fine-tuning of batchnorm parameters can mitigate the impact of non-idealities during inference.

    \item We perform a holistic evaluation of our proposed approach using a framework to capture the impact of different types of crossbar non-idealities affecting inference- stochastic read noise, temporal drift and parasitic resistances.

    \item We consider state-of-the-art VGG16 \& ResNet-18 DNNs trained using CIFAR10 \& TinyImagenet datasets, respectively, for our experiments. We perform ablations with memristive devices (FeFET, PCM and RRAM) to quantify the non-linear shift (using Principal Component Analysis) introduced by the crossbar non-idealities on the convolution weights. Based on this, we suggest a nearly \textit{training-less} non-ideality mitigation technique to preserve DNN accuracy on hardware.

    \item We evaluate the advantages offered by our proposed approach in terms of training energy- \& memory-savings on hardware that supports both online training and inference. 
    
\end{itemize}

\section{Background}
\label{sec:back}
\vspace{-1mm}
\subsection{Memristive crossbars and non-idealities}
\vspace{-1mm}

\begin{wrapfigure}{l}{0.2\textwidth}
\includegraphics[width=0.2\textwidth]{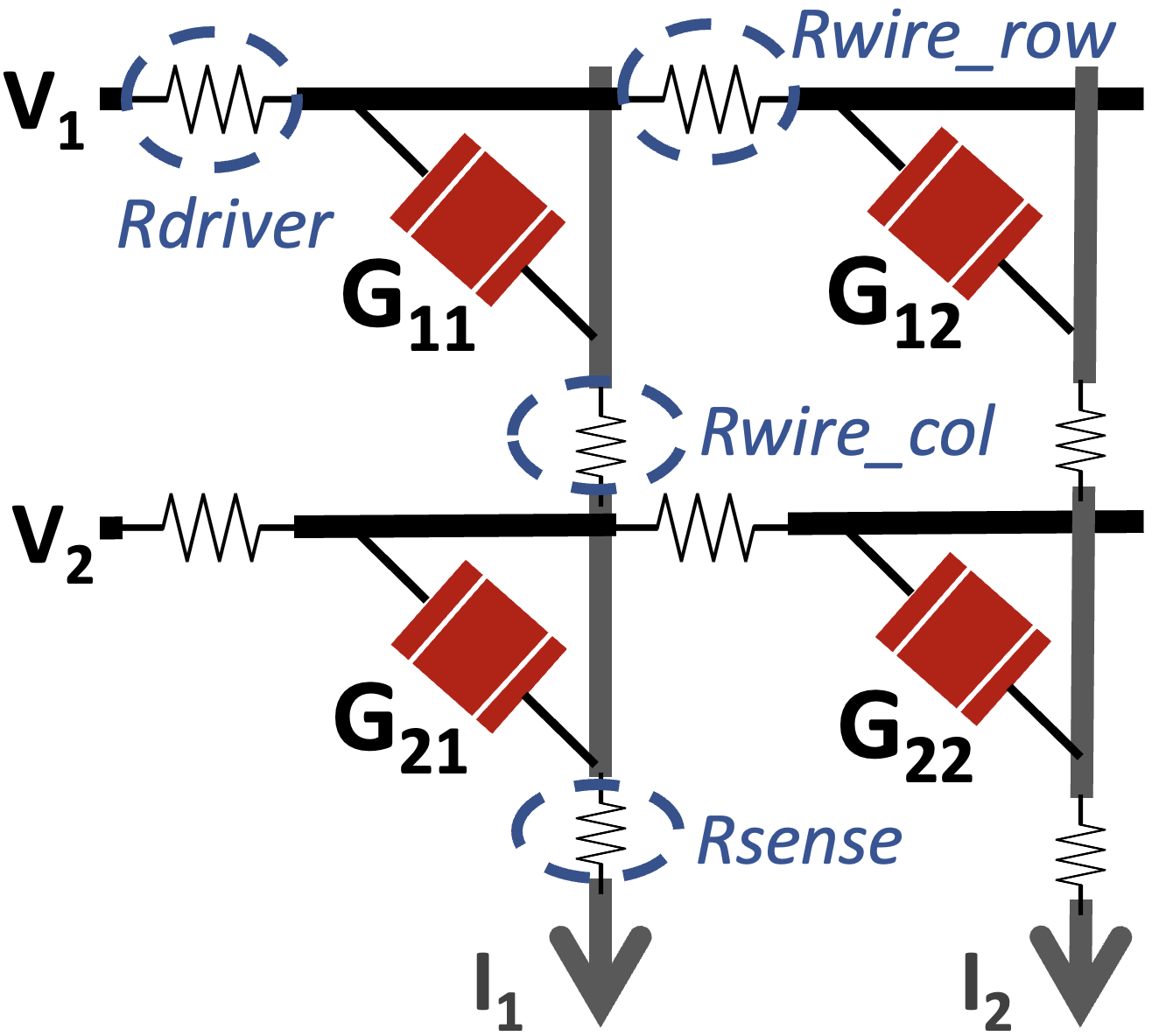}
  
\caption{A 2$\times$2 memristive crossbar with parasitic resistive non-idealities annotated. Note, the device conductances $G_{ij}$s are susceptible to read and drift noise.}
\label{xbar}
\vspace{-4mm}
\end{wrapfigure}

Analog crossbars consist of 2D arrays of memristive devices interfaced with Digital-to-Analog Converters (DACs), Analog-to-Digital Converters (ADCs) and a programming circuit. The synaptic devices are programmed to a particular value of conductance (between $G_{MIN}$ and $G_{MAX}$) during inference. For emulating Multiply-and-Accumulate (MAC) operations in DNNs, the activations are fed in as analog voltages $V_i$ to each row using DACs and weights are programmed as synaptic device conductances ($G_{ij}$) as shown in Fig. \ref{xbar} \cite{jain2020rxnn, bhattacharjee2021neat, sebastian2020memory}. For an ideal crossbar array, the voltages interact with the device conductances and produce a current (governed by Ohm's Law) during inference. Consequently, by Kirchoff's current law, the net output current sensed by ADCs at each column $j$ or Bit-line (BL) is the sum of currents through each device, \textit{i.e.} $I_{j(ideal)} = \Sigma_{i}^{}{G_{ij} * V_i}$.
\vspace{-3mm}

\textbf{Impact of non-idealities:} In reality, the analog nature of the computation leads to various hardware noise or non-idealities, such as device-level variations of the memristors and interconnect parasitic resistances in the crossbars \cite{jain2020rxnn, sun2019impact, chakraborty2020geniex, antolini2023combined} (see Fig. \ref{xbar}). Consequently, the net output current sensed at each column $j$ in a non-ideal scenario becomes $I_{j(non-ideal)} = \Sigma_{i}^{}{G_{ij}' * V_i}$, which deviates from its ideal value. This manifests as accuracy degradation for DNNs mapped onto analog crossbars. This work deals with three types of non-idealities during inference- stochastic read, temporal drift and resistive parasitic non-idealities. 

\begin{figure*}[t]
    \centering
    \includegraphics[width=\linewidth]{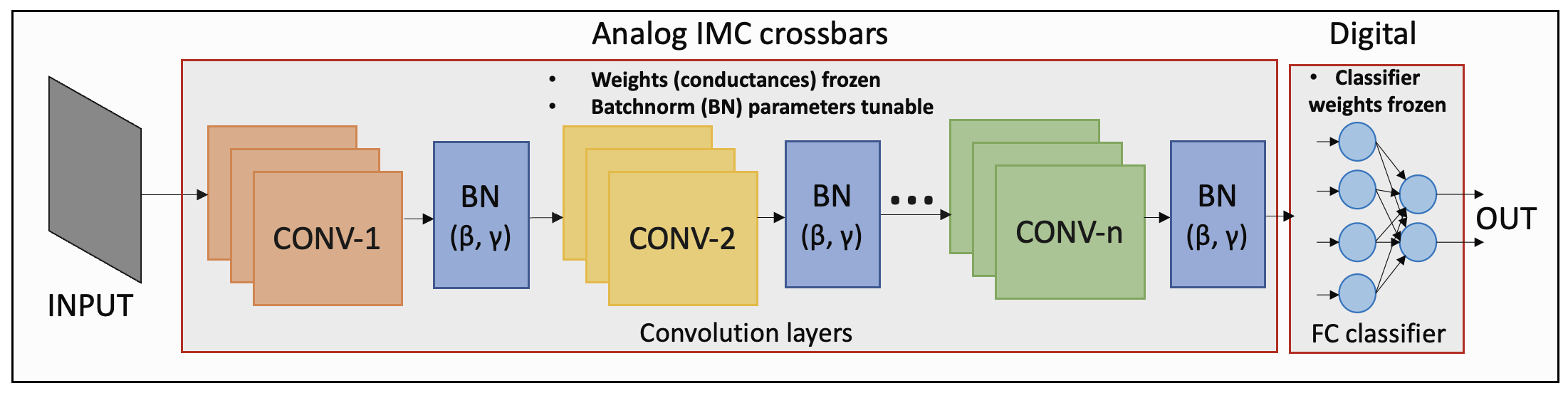}
   
    \caption{An illustration of the implementation of a DNN on a hybrid analog-digital hardware architecture. The convolution layers are implemented on analog crossbars, while the BN and FC classifier layers are implemented using digital CMOS technology.}
    \label{implementation}
     
\end{figure*}

Memristive devices exhibit device-to-device variations in their conductances $G$ that constitute the stochastic read noise $\tilde{n} \propto \mathcal{N}(0,\sigma^{2})$, where $\sigma$ signifies the standard-deviation of noise \cite{sun2019impact}. The noisy conductance $G'$ can be modelled as:

\vspace{-4mm}
\small
\begin{equation} G'=G+\tilde{n}.
\label{eq:read-noise}
\end{equation}
\normalsize

Memristive devices are also susceptible to temporal drift in their conductances and with time, their programmed conductance decreases and approaches higher resistance states \cite{sun2019impact}. The effect of temporal drift can be modelled using the equation:

\vspace{-3mm}
\small
\begin{equation} G'=G_0*(\frac{T}{T_0})^{-\nu}.
\label{eq:drift-noise}
\end{equation}
\normalsize

Here, $T$ denotes time elapsed since programming the device to conductance $G_0$ at time $T_0$ ($T_0$ is assumed to be 1s) and $\nu$ denotes the drift coefficient of the memristive device. Higher the value of $\nu$, poorer the \textit{retention} of the device during inference. Note, in our experiments $T$ denotes the time since the programming of the devices at which the inference accuracy is measured.

Fig. \ref{xbar} indicates the various resistive non-idealities, \textit{viz.} $Rdriver$, $Rwire\_row$, $Rwire\_col$ and $Rsense$, prevalant in crossbars \cite{jain2020rxnn, bhattacharjee2021neat}. Prior works have modelled the impact of resistive non-idealities by using linear algebraic operations and circuit laws to transform the ideal memristive conductances to non-ideal conductances  \cite{jain2020rxnn, roy2021txsim}.
\vspace{-3mm}
\subsection{Batch Normalization in DNNs}
\vspace{-1mm}
Today, top-performing deep convolutional neural networks inevitably use batchnorm (BN) layers. The BN layers minimize the internal co-variate shifting, which is a common problem while training DNNs, thereby achieving faster convergence and higher performance \cite{ioffe2015batch, szegedy2016rethinking}.  Let $X~\epsilon~\mathbb{R}^{n\times p}$ denote the input to a BN layer, where $n$ is the batch size and $p$ is the feature dimension. Batch normalization involves two statistical components, mean ($\mathbb{E}[X_j]$) \& variance ($Var(X_j)$), for standardization of each feature in a mini-batch, and thereafter learning a common slope and bias corresponding to each mini-batch, using the training data. The BN layer outputs a feature $j~\epsilon~\{1,...,p\}$ into:

\vspace{-2mm}
\small
\begin{equation} {x'}_j = \frac{x_j-\mathbb{E}[X_j]}{\sqrt{Var(X_j)}}; \qquad y_j = \gamma_j {x'}_j + \beta_j.
\label{eq:bn}
\end{equation}
\normalsize

Here, ${x}_j$ and $y_j$ denote the input and output scalars of one neuron response in one data sample, $X_j$ denotes the $j^{th}$ column of the input data and $\gamma_j$ and $\beta_j$ refer to the learnable/tunable batchnorm parameters. Batch normalization ensures that the input distribution across different mini-batches is not altered. 

\textbf{Batchnorm adaptation \& Batchnorm fine-tuning: }From equation (\ref{eq:bn}), we find that batch normalization involves a linear transformation on $x_j$ that results in $y_j$. When DNNs are deployed on crossbars, $x_j$s are generated as a result of dot-product operations along the crossbar columns. The crossbar non-idealities lead to a deviation of $x_j$ from its ideal value resulting in noisy crossbar activations, which can be corrected by noise-aware BN adaptation. BN adaptation updates the average mean \& variance in the BN layers. Specifically, a number of training image samples are forwarded through the DNN, and the moving average \& variance of BN layers are adapted with respect to non-ideal crossbar activations (while keeping all other learnable parameters frozen) \cite{tsai2020robust, bhattacharjee2022examining}. However, we will see in this work that if the degree of non-linearity in $x_j$ introduced by the crossbar non-idealities is too large, simple BN adaptation can fall short as a non-linearity mitigation strategy. To this end, we propose noise-aware fine-tuning of the $\gamma_j$ and $\beta_j$ parameters in the BN layers, while keeping all the weights frozen. We term this approach as \textit{batchnorm fine-tuning}. 
\vspace{-3mm}

\section{Hardware Implementation and Methodology}
\label{sec:method}
\vspace{-1mm}

As shown in Fig. \ref{implementation}, we consider DNNs implemented on a hybrid analog-digital hardware architecture, wherein all the convolutional layers are mapped onto analog crossbar-arrays using low-precision memristive devices in a weight-stationary manner, and the fully-connected (FC) classifier layer is implemented using digital CMOS technology at full floating-point precision \cite{ueyoshi2022diana}. Note that the BN, pooling and activation units are also implemented using digital CMOS circuits. Thus, the analog non-idealities only impact the weights of convolution layers in the DNNs. In our experiments (Section \ref{sec:expt}), we keep all the weights (or conductances) in the crossbars frozen, and only the batchnorm parameters ($\beta$ and $\gamma$) are trainable. 

\begin{figure}[t]
    \centering
    \includegraphics[width=\linewidth]{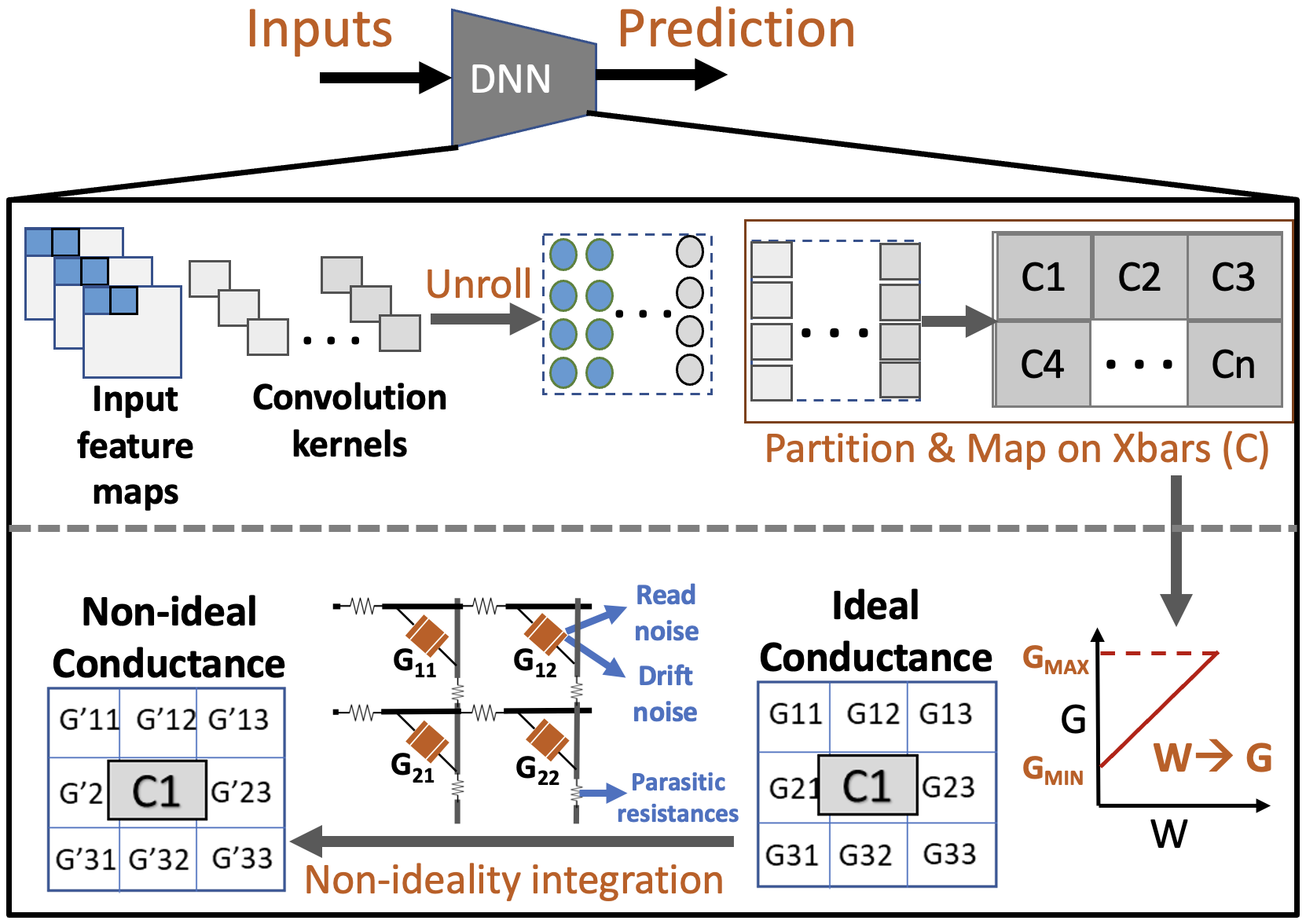}
   
    \caption{An illustration of our non-ideality integration framework to simulate the impact of crossbar non-idealities on DNN weights.}
    \label{framework}
      
\end{figure}

\begin{table}[t]
\caption{Table showing the properties of the memristive devices. Here, $G$ denotes the memristive conductance in $\mu S$.}
\label{tab:devices}
\resizebox{\linewidth}{!}{%
\begin{tabular}{|c|c|c|c|c|}
\hline
\textbf{Device} &
  \textbf{\begin{tabular}[c]{@{}c@{}}Precision \\ (bits)\end{tabular}} &
  \textbf{\begin{tabular}[c]{@{}c@{}}$R_{ON}$, \\ ON/OFF ratio\end{tabular}} &
  \textbf{\begin{tabular}[c]{@{}c@{}}Drift \\ coefficient \\ ($\nu$)\end{tabular}} &
  \textbf{\begin{tabular}[c]{@{}c@{}}Read noise \\ ($\sigma$)\end{tabular}} \\ \hline
FeFET \cite{byun2022recent}  & 4 & \begin{tabular}[c]{@{}c@{}}222.22 k$\Omega$,\\ 100\end{tabular} & 0.1  & 0.05  \\ \hline
PCM-i \cite{nandakumar2018phase}  & 4 & \begin{tabular}[c]{@{}c@{}}250 k$\Omega$,\\ 40\end{tabular}     & 0.04 & $\sigma($G$) = 0.03$G$~+~0.13$  \\ \hline
PCM-ii \cite{nandakumar2018phase} & 4 & \begin{tabular}[c]{@{}c@{}}125 k$\Omega$,\\ 80\end{tabular}     & 0.04 & $\sigma($G$) = 0.03$G$~+~0.13$  \\ \hline
RRAM \cite{hajri2019rram}   & 4 & \begin{tabular}[c]{@{}c@{}}50 k$\Omega$,\\ 10\end{tabular}      & 0.04 & 0.1 \\ \hline
\end{tabular}%
}

\end{table}

\textbf{Non-ideality integration framework:} In our experiments, we simulate the impact of various analog non-idealities on the weights of the convolution layers which are deployed on crossbars. Fig. \ref{framework} describes our non-ideality integration framework for the convolution weights. The entire framework is written in Python. It involves a Python wrapper that reshapes the 4D convolution weight-tensors of each layer of a pre-trained DNN into 2D weight-matrices ($W$) consisting of the ideal weights. Thereafter, the matrices are partitioned into multiple crossbar instances (of a given size), followed by the conversion of the ideal weights into synaptic conductances programmed in the memristive devices with a particular precision. The properties of the different devices used in our experiments are specified in Table \ref{tab:devices}. To the ideal conductances, we include the impact of various types of non-idealities (specified in Section \ref{sec:back}). Here, we model stochastic read and temporal drift noise using equations (\ref{eq:read-noise}) \& (\ref{eq:drift-noise}), respectively. The impact of the resistive parasitic non-idealities in the crossbars is modelled using circuit laws and linear algebraic operations written in Python \cite{jain2020rxnn, bhattacharjee2022examining}. Finally, we transform the non-ideal conductances into non-ideal weights which are then integrated into the original Pytorch based DNN model for inference or crossbar noise-aware fine-tuning of the BN parameters. 
\vspace{-7mm}
\section{Experiments and Results}
\label{sec:expt}
\vspace{-1mm}
\textbf{Experimental setup:} We use VGG16 and ResNet-18 models pre-trained at full-precision with CIFAR10 (test accuracy = \textbf{88.62 \%}) and TinyImagenet (test accuracy = \textbf{51.80 \%}) datasets, respectively. 
CIFAR10 has RGB images (50,000 training and 10,000 testing) of size 32$\times$32 belonging to 10 classes. TinyImagenet is a more complex dataset with RGB images (100,000 training and 10,000 testing) of size 64$\times$64 belonging to 200 classes. 
The DNNs are deployed on analog crossbar platforms built using memristive devices specified in Table \ref{tab:devices}. Note, we realize 8-bit weights using the 4-bit memristive devices via bit-slicing \cite{peng2020dnn+} and the input activations to the crossbars are assumed to be 8-bits. Also, the resistive non-idealities in the crossbars (see Fig. \ref{xbar}) are as follows: $Rdriver = 1 k\Omega$, $Rwire\_row = 5 \Omega$, $Rwire\_col = 10 \Omega$ and $Rsense = 1 k\Omega$ \cite{bhattacharjee2021neat, bhattacharjee2022examining}. Unless otherwise stated, we consider a crossbar size of 64$\times$64 in our experiments.

\subsection{Impact of stochastic read noise} 
\vspace{-1mm}
\begin{figure}[htbp]
    \centering
    \includegraphics[width=.9\linewidth]{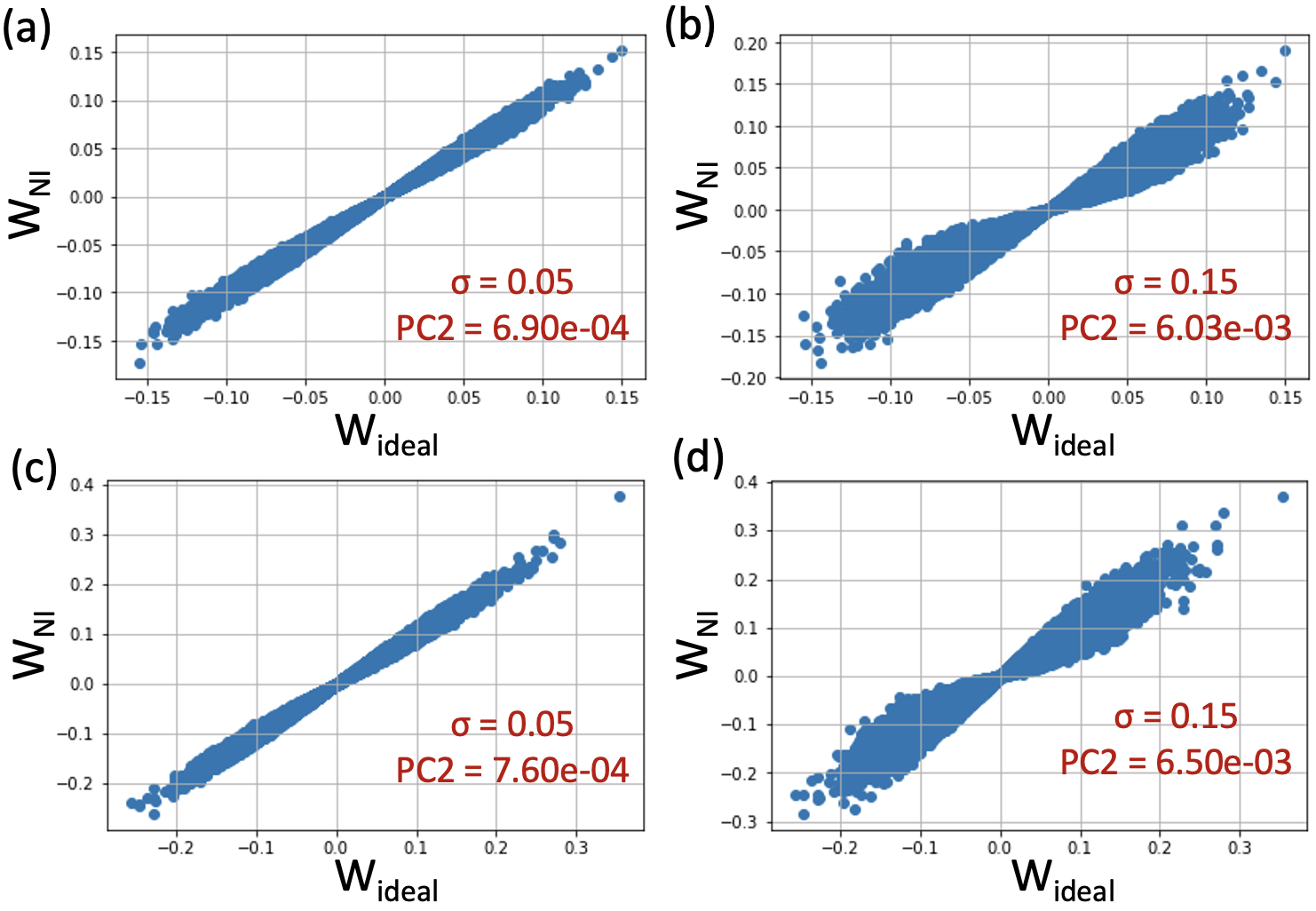}
   
    \caption{Plot of the distribution of $W_{NI}$ against $W_{ideal}$ in the- (a) CONV-5 layer of the VGG16/CIFAR10 model for $\sigma=0.05$. (b) CONV-5 layer of the VGG16/CIFAR10 model for $\sigma=0.15$. (c) CONV-2 layer in the last residual block of the ResNet-18/TinyImagenet model for $\sigma=0.05$. (d) CONV-2 layer in the last residual block of the ResNet-18/TinyImagenet model for $\sigma=0.15$.}
    \label{r1}
  
\end{figure}

We map the VGG16 \& ResNet-18 DNNs onto FeFET crossbars. We vary the read noise of the device from $\sigma=0.05$ to $\sigma=0.35$ and assume all other crossbar non-idealities to be absent. In Fig. \ref{r1}(a-d), we plot the distribution of non-ideal weights ($W_{NI}$) with respect to the ideal software weights ($W_{ideal}$) in the CONV-5 layer of the VGG16 model and the CONV-2 layer in the last residual block of the ResNet-18 model, for $\sigma=0.05$ and $\sigma=0.15$. We visually observe that the distribution assumes the shape of a \textit{dumb-bell} with the degree of non-linear shift (width of the dumb-bell) increasing due to rising noise from $\sigma=0.05$ to $\sigma=0.15$. To quantify the degree of non-linear shift, we perform a Principal Component Analysis (PCA) of the two-dimensional distribution between $W_{NI}$ and $W_{ideal}$ and compute the proportion of variance due the first and the second principal components (PC1 and PC2, respectively). \begin{wrapfigure}{l}{0.24\textwidth}
\includegraphics[width=0.24\textwidth]{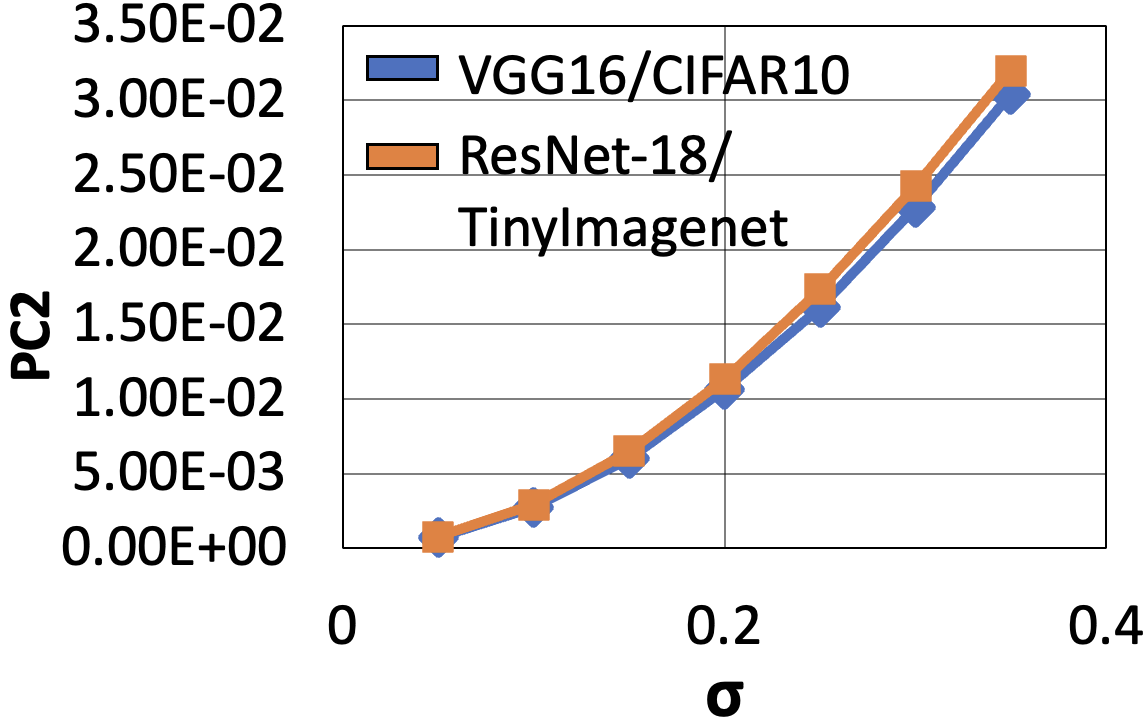}
  
\caption{Plotting PC2 for the CONV-5 layer of the VGG16/CIFAR10 model (blue) and the CONV-2 layer in the last residual block of the ResNet-18/TinyImagenet model against $\sigma$ varying from 0.05 to 0.35.}
\label{r2}
\vspace{-6mm}
\end{wrapfigure} Ideally, when $W_{NI}=W_{ideal}$, $PC1 = 1.0$ \& $PC2 = 0.0$ which implies that the data is distributed only along the line $W_{NI}=W_{ideal}$ and there is no distribution shift orthogonal to it. However, we observe $PC1<1.0$ and $PC2>0.0$ when there is a non-linear shift due to crossbar noise. As shown in Fig. \ref{r2}, as the amount of read noise increases from $\sigma=0.05$ to $\sigma=0.35$, the value of PC2 increases indicating higher non-linear shift in the distribution between $W_{NI}$ and $W_{ideal}$. 

\begin{figure}[t]
    \centering
    \includegraphics[width=0.9\linewidth]{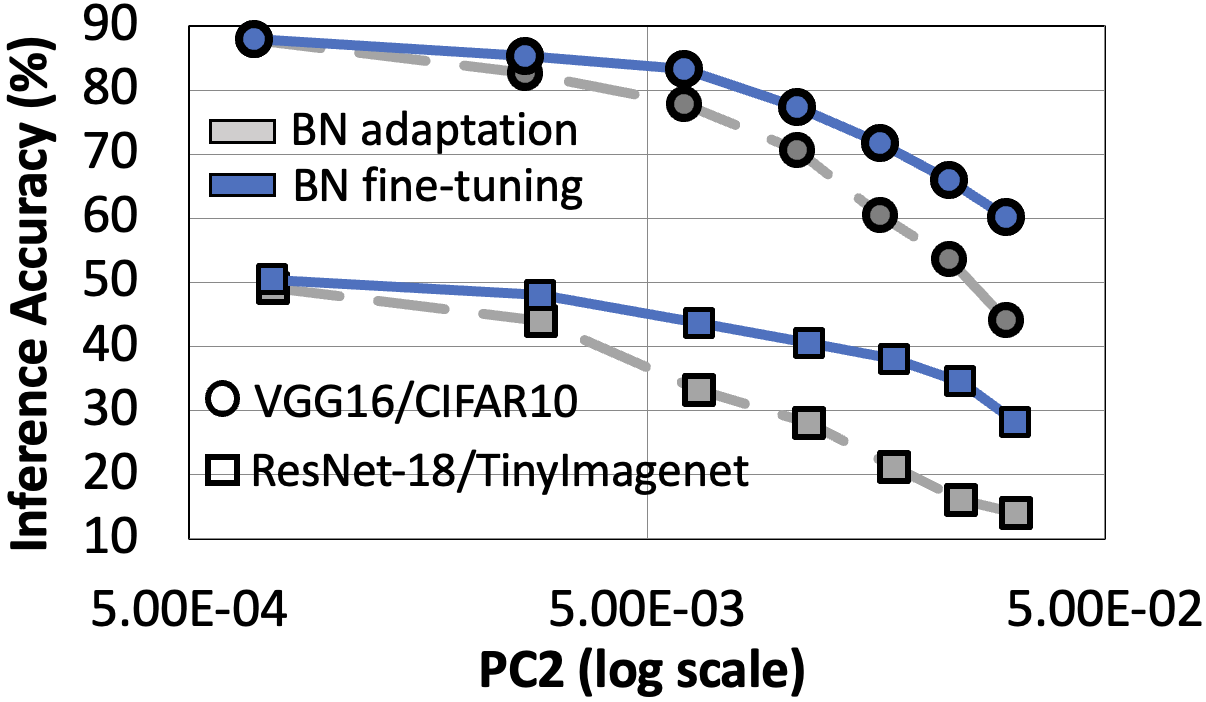}
   
    \caption{Plot of non-ideal accuracy against PC2 ($\sigma=0.05-0.35$) for VGG16/CIFAR10 (circles) and ResNet-18/TinyImagenet (squares).}
    \label{r3}
     
\end{figure}

Now in Fig. \ref{r3}, we plot the non-ideal inference accuracy of the VGG16/CIFAR10 \& ResNet-18/TinyImagenet models with respect to the PC2 values for $\sigma=0.05-0.35$ after carrying out crossbar-aware batchnorm adaptation (grey curves). For the VGG16/CIFAR10 model at a lower value of read noise $\sigma=0.05$, the value of PC2 is small ($PC2 = 6.90e-04$) and the non-ideal accuracy is preserved at 87.78 \% with batchnorm adaptation. We find that beyond $\sigma=0.1$ ($PC2 = 2.71e-03$), the non-ideal accuracy cannot be well-preserved as the non-linear shift in the distribution between $W_{NI}$ and $W_{ideal}$ becomes too large to be compensated by simple batchnorm adaptation. At $PC2 = 6.03e-03$ for $\sigma=0.15$, the non-ideal accuracy stands at 77.89 \% even with batchnorm adaptation. Similar trends are also seen for the ResNet-18/TinyImagenet model, where the non-ideal accuracy with batchnorm adaptation dramatically decreases at $\sigma=0.1$ ($PC2 = 2.92e-03$). 

\textbf{Fine-tuning batchnorm parameters ($\beta$ \& $\gamma$):} Since, the BN layers in both VGG16 and ResNet-18 models are implemented in digital CMOS technology at full-precision and assumed to be free from non-idealities, it incurs very low overhead for bathcnorm fine-tuning on hardware. Hence, we only fine-tune the batchnorm parameters ($\beta$ \& $\gamma$) for 5 epochs. Note, batchnorm adaptation is implicit within batchnorm fine-tuning. For all experiments, the fine-tuning occurs with an initial learning rate is 0.01 that is reduced by 5 times at every second epoch. With this approach, we find in Fig. \ref{r3} (blue curves) that the non-ideal accuracy of the VGG16/CIFAR10 model on FeFET-based crossbars can be fairly preserved till $PC2 < 1.0e-02$ ($\sigma=0.15-0.2$), beyond which the accuracy begins to dramatically decrease as the read noise and hence, the value of PC2 increases. Similar trends can be seen for the ResNet-18/TinyImagenet model, where the read noise-sensitivity is higher and the non-ideal accuracy can be fairly preserved till $\sigma=0.1$ ($PC2 = 2.92e-03$). 

\begin{figure*}[t]
    \centering
    \includegraphics[width=0.85\linewidth]{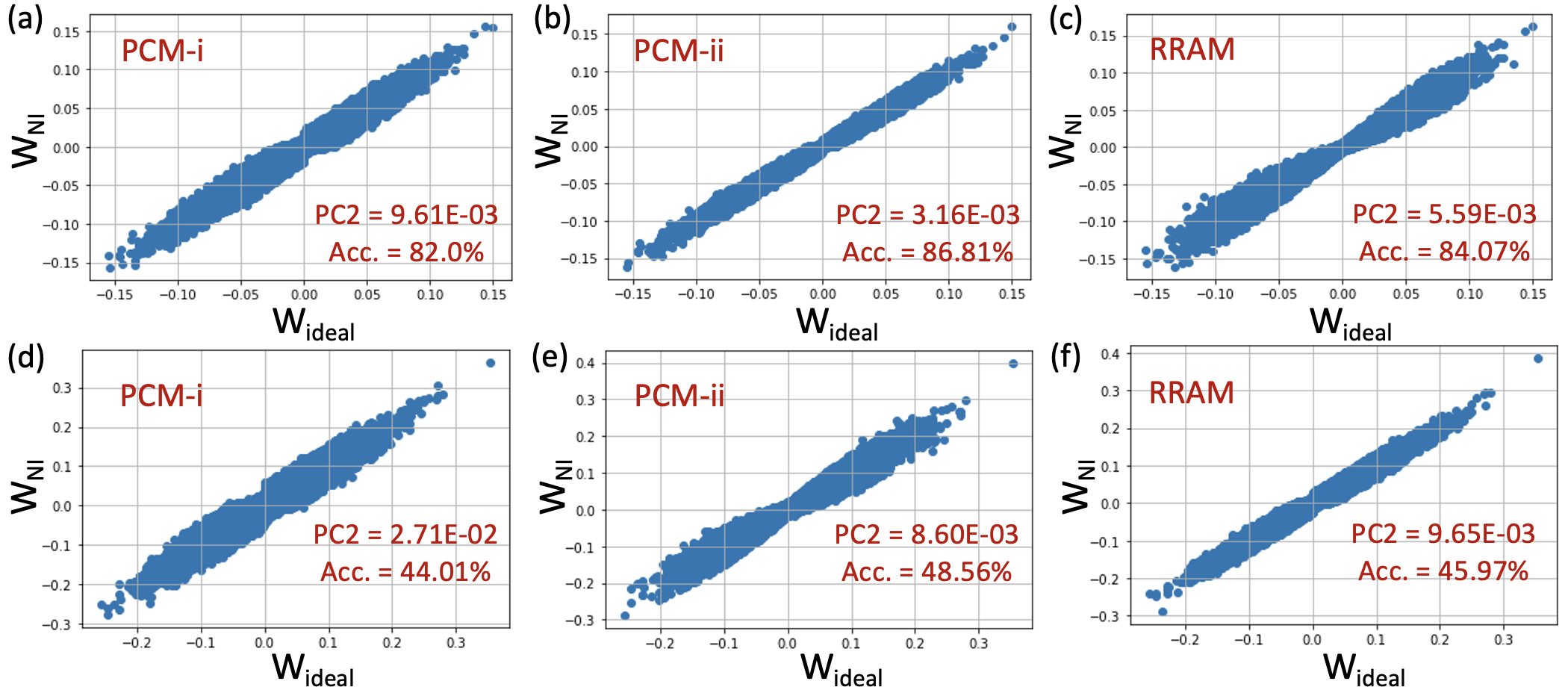}
   
    \caption{(a-c) Plot of the distribution of $W_{NI}$ with respect to that of $W_{ideal}$ in the CONV-5 layer of the VGG16/CIFAR10 model for mapping onto- (a) PCM-i, (b) PCM-ii and (c) RRAM crossbars. (d-f) Plot of the distribution of $W_{NI}$ with respect to that of $W_{ideal}$ in the CONV-2 layer in the last residual block of the ResNet-18/TinyImagenet model for mapping onto- (d) PCM-i, (e) PCM-ii and (f) RRAM crossbars. Note, the accuracies annotated are the non-ideal accuracies of the DNNs post noise-aware batchnorm fine-tuning.}
    \label{r4}
      
\end{figure*}

\textbf{Impact of device ON/OFF ratio:} We show results for the VGG16/CIFAR10 DNN mapped onto PCM-i crossbars (ON/OFF ratio = 40). Fig. \ref{r4}(a) shows the distribution between $W_{NI}$ and $W_{ideal}$ for the CONV-5 layer. The value of PC2 is $9.61e-03$, and it achieves 82.0 \% accuracy upon BN fine-tuning. However, with simple BN adaptation, its accuracy is 79.85 \% due to a higher value of PC2. Interestingly, we observe that if the VGG16/CIFAR10 DNN is mapped using PCM-ii devices (ON/OFF ratio = 80) as shown in Fig. \ref{r4}(b), the value of PC2 reduces to $3.16e-03$ and the accuracy upon BN fine-tuning increases to 86.81 \%. The inference accuracy with simple BN adaptation drops to 85.44 \%. This is because with increased ON/OFF ratio, the interval between two programmable device conductance levels increases, and thus, the noise-sensitivity of the device reduces which manifests as reduction in the value of PC2. Now, let us compare with a VGG16/CIFAR10 DNN mapped onto crossbars based on RRAM devices (ON/OFF ratio = 10), as shown in Fig. \ref{r4}(c). Although RRAM ($\sigma=0.1$) has a lower impact of read noise, it also has a very limited ON/OFF ratio which makes the device sensitive to noise and thus, its $PC2 = 5.59e-03$ and the accuracy upon BN fine-tuning is 84.07 \%. With simple BN adaptation, the accuracy drops to 80.40 \%. Similar trends can also be seen for the ResNet-18/TinyImagenet model when deployed on crossbars based on PCM-i, PCM-ii and RRAM devices, as shown in Fig. \ref{r4}(d-f).

\subsection{Impact of temporal drift noise}
\vspace{-1mm}
\begin{figure}[t]
    \centering
    \includegraphics[width=.9\linewidth]{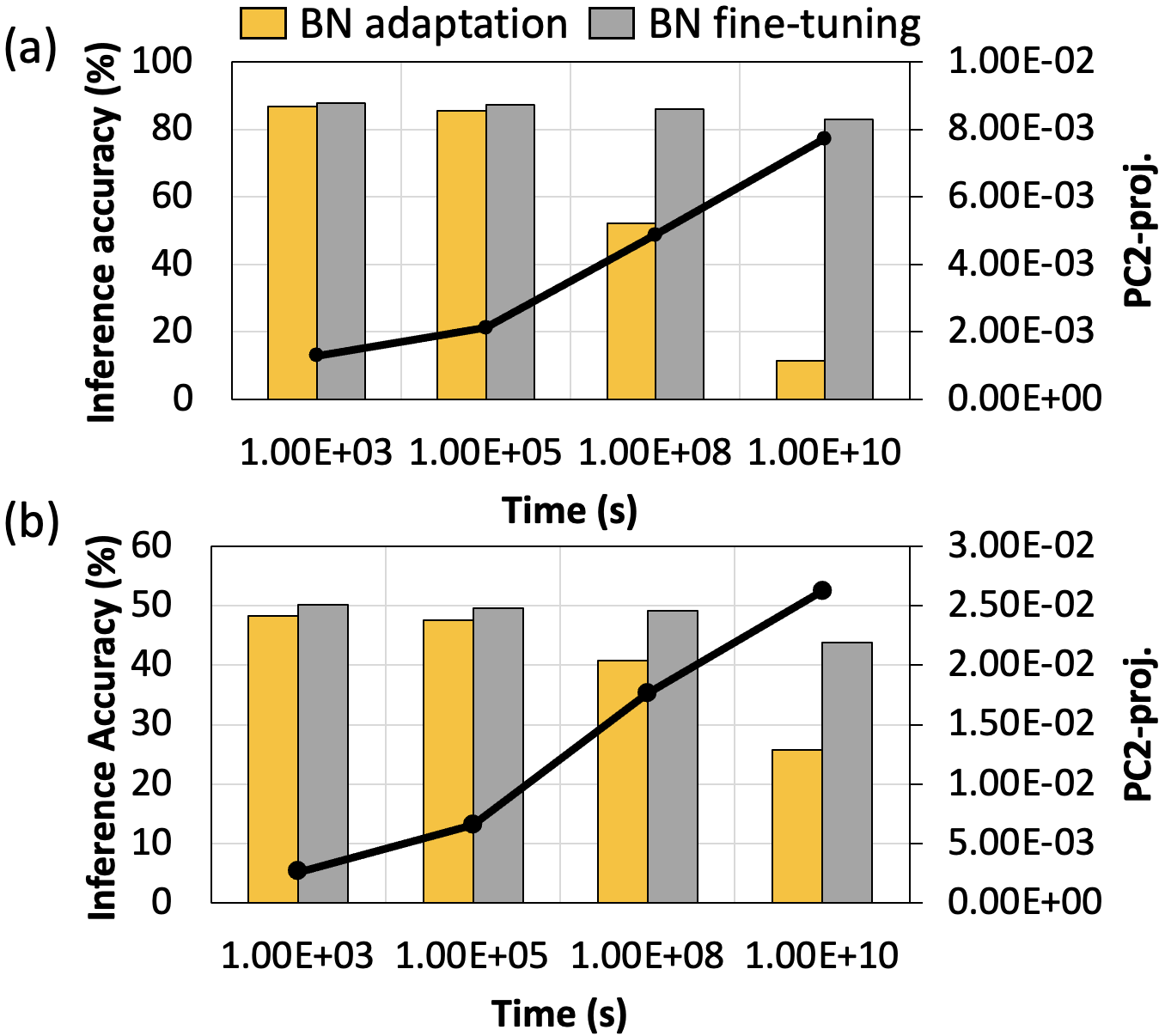}
   
    \caption{(Primary y-axis) Bar plot of inference accuracy versus time $T$ for- (a) VGG16/CIFAR10 and (b) ResNet-18/TinyImagenet models. (Secondary y-axis) Line plot of PC2-proj. versus time $T$ for- (a) CONV-5 layer of the VGG16/CIFAR10 and (b) CONV-2 layer in the last residual block of the ResNet-18/TinyImagenet.}
    \label{r5}
    \vspace{-5mm}
\end{figure}

The inference accuracy on crossbars gradually decreases with time $T$ due to temporal drift of the conductances in the memristive devices (see Section \ref{sec:back}). FeFET devices typically have higher temporal drift coefficients (see Table \ref{tab:devices}) and so, DNNs deployed on FeFET crossbars show lower \textit{retention} of accuracy. Drift noise-aware BN adaptation can maintain the accuracy of crossbar-mapped DNNs till a retention time $T$ beyond which the accuracy dramatically decreases \cite{joshi2020accurate}. From Fig. \ref{r5}(a-b), we find the retention time for VGG16/CIFAR10 \& ResNet-18/TinyImagenet models on FeFET crossbars with simple BN adaptation to be $\sim10^4-10^5s$. 
Beyond that, the non-ideal accuracy dramatically decreases and at $T=10^{10}s$, the VGG16 and ResNet-18 models, respectively, suffer a 76.5 \% and 26.1 \% drop in accuracy with respect to their initial inference accuracies. 

\begin{wrapfigure}{l}{0.23\textwidth}
\includegraphics[width=0.23\textwidth]{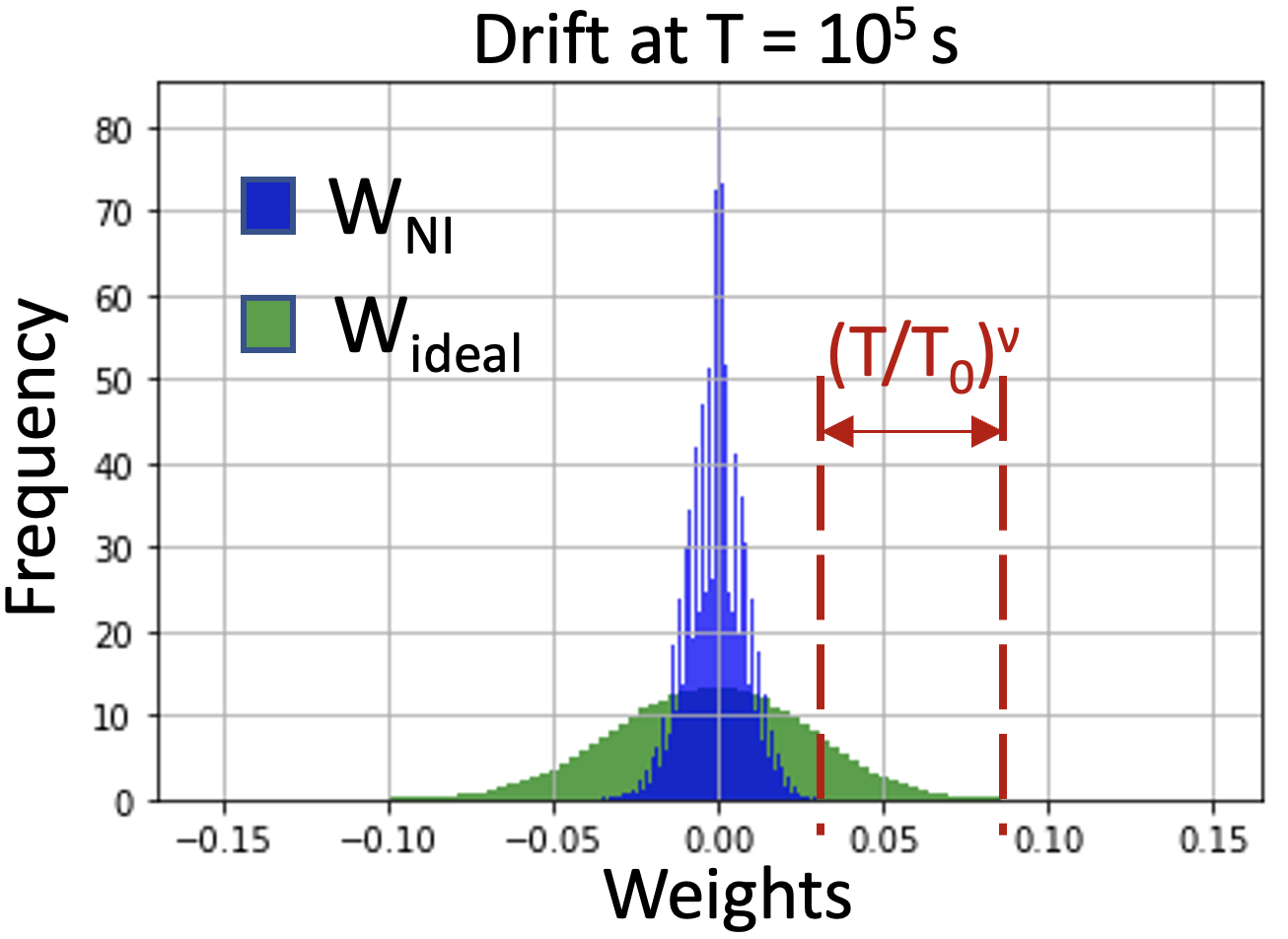}
  
\caption{Histogram plot for the distribution of $W_{ideal}$ (green) and $W_{NI}$ (blue) for the CONV-5 layer of VGG16/CIFAR10 at $T=10^{5}s$.}
\label{r6}
 \vspace{-6mm}
\end{wrapfigure}

Fig. \ref{r6} shows the histogram of the distribution of weights $W_{ideal}$ (green) and $W_{NI}$ (blue) at $T=10^{5}s$, for the CONV-5 layer of the VGG16/CIFAR10 model. We find that drift noise results in constriction of the range of non-ideal weights (or conductances) by a factor of $(T/T_0)^\nu$. This is unlike other types of crossbar noise, \textit{e.g.} stochastic read noise, where the ranges of $W_{ideal}$ and $W_{NI}$ are similar (see Fig. \ref{r1}). Thus, in case of drift noise at time $T$, we will project the PC2 values for the distribution between $W_{ideal}$ and $W_{NI}$ by scaling up the measured PC2 with $(T/T_0)^\nu$. We use PC2-proj. to denote this projected value of PC2. Fig. \ref{r5}(a-b) shows that as we increase the value of $T$, the value of PC2-proj. increases, due to a higher drift noise. We find that beyond $\sim10^4-10^5s$ (\textit{i.e}, PC2-proj. > 0.002 for VGG16 and PC2-proj. > 0.006 for ResNet-18), the accuracies cannot be preserved using simple BN adaptation. With BN fine-tuning, the accuracies can be well-preserved at very high values of $T$ (or PC2-proj.) and the accuracy losses at $T=10^{10}s$ are reduced to 5.0 \% and 8.0 \% for the VGG16 and ResNet-18 models, respectively.

\vspace{-3mm}
\subsection{Impact of resistive parasitic non-idealities}
\vspace{-1mm}   
\begin{figure}[t]
    \centering
    \includegraphics[width=.9\linewidth]{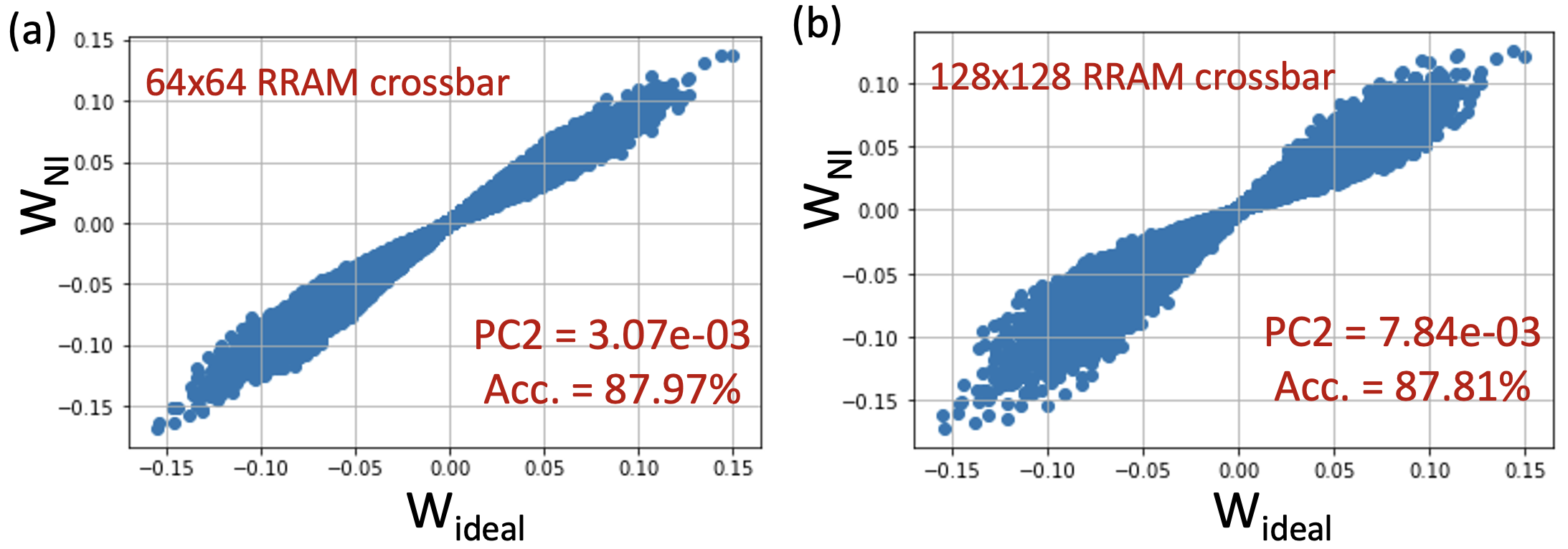}
   
    \caption{Plot of the distribution of $W_{NI}$ with $W_{ideal}$ in the CONV-5 layer of the VGG16/CIFAR10 model for mapping on RRAM-based crossbars of size- (a) 64$\times$64 and (b) 128$\times$128, having only resistive parasitic non-idealities. Note, the accuracies annotated are the non-ideal accuracies of the DNNs post noise-aware batchnorm fine-tuning.}
    \label{r7}

\end{figure}

We consider the impact of resistive parasitic non-idealities alone on the weights of VGG16/CIFAR10 on RRAM crossbars of size 64$\times$64 and 128$\times$128, without read or drift noise. Prior works \cite{chakraborty2020geniex, bhattacharjee2021neat, rasch2023hardware} show that lower the value of $R_{ON}$, greater the impact of resistive non-idealities on the DNN weights, leading to significant accuracy losses on hardware. From Table \ref{tab:devices}, RRAM has the lowest value of $R_{ON}$ and ON/OFF ratio among all the devices and thus, DNN weights mapped on RRAM crossbars will be maximally impacted by the resistive non-idealities. Also, larger crossbars entail greater impact of resistive crossbar non-idealities. 

Fig. \ref{r7}(a-b) show the distribution between $W_{NI}$ and $W_{ideal}$ for the CONV-5 layer for crossbars of size 64$\times$64 and 128$\times$128, respectively. For 64$\times$64 (128$\times$128) crossbars, the value of PC2 is $3.07e-03$ ($7.84e-03$), and the DNN inference accuracy on hardware is reduced to 74.37 \% (32.9 \%). However, at this regime of PC2 ($PC2<1.0e-02$), most of the inference accuracy can be recovered with simple crossbar-aware BN adaptation, and the accuracy is increased to 86.31 \% (83.03 \%). On carrying out BN fine-tuning, the inference accuracy is 87.97 \% (87.81 \%) which has been annotated in Fig. \ref{r7}. This shows that resistive non-idealities, although can disrupt a DNN's performance on crossbars, can be totally mitigated with the help of BN fine-tuning, without the need for parasitic noise-aware re-training of weights as proposed in \cite{roy2021txsim, negi2021nax, rasch2023hardware}. Now, let us superimpose the stochastic read noise of $\sigma=0.1$ for the RRAM devices over the resistive parasitic non-idealities. The inference accuracy on hardware is 62.28 \% (27.86 \%), which on BN fine-tuning is raised to 83.22 \% (81.59 \%).

\section{Hardware Advantages}
\vspace{-1mm}

Consider VGG16 and ResNet-18 DNNs with their convolution layers deployed on 64$\times$64 analog RRAM crossbars, supporting online training and inference \cite{peng2020dnn+}. Note that the BN, pooling \& activation units as well as the classifier are implemented using digital CMOS circuits (see Fig. \ref{implementation}). Since weights are frozen, there is no compute \& storage requirement for the weight-gradients and weight update (writing into the RRAM devices) during training. However, we compute and store the activations and their gradients to calculate $\nabla_\gamma \Lagr$ \& $\nabla_\beta \Lagr$  for updating $\gamma$ \& $\beta$. Note, $\nabla_x f$ denotes gradient of $f$ with respect to $x$ and $\Lagr$ is the cross-entropy loss. However, the memory overhead (in MB) for storing the weight-gradients is much greater than the activation-gradients. \begin{wrapfigure}{l}{0.24\textwidth}
\includegraphics[width=0.24\textwidth]{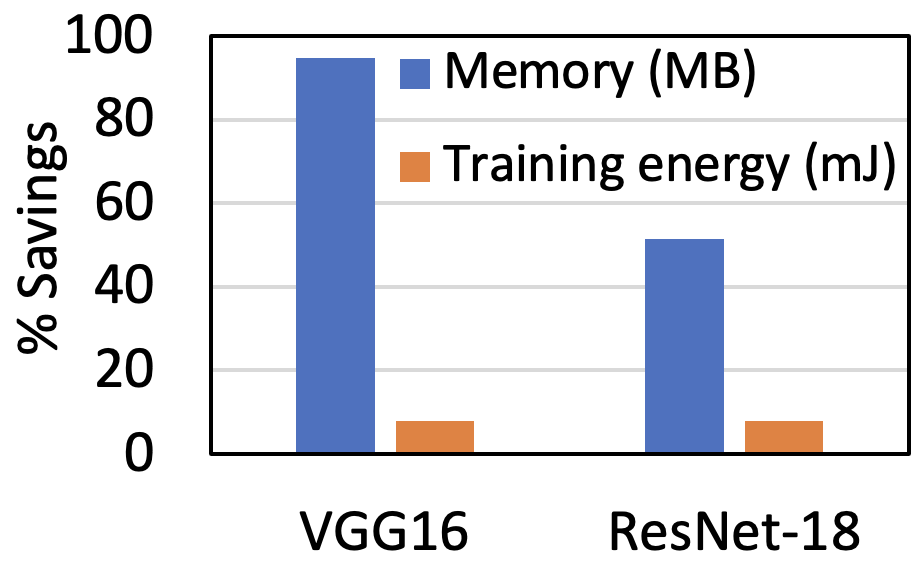}
  
\caption{Bar plot showing percentage savings in training energy (mJ) and memory (MB) on hardware with our proposed approach of crossbar noise-aware batchnorm fine-tuning keeping DNN weights frozen.}
\label{hb1}

\end{wrapfigure}Consequently, there is a huge memory-saving by not storing the weight-gradients during backpropagation. Likewise, since the RRAM devices will not require re-programming, we reduce the training energy on hardware. Fig. \ref{hb1} shows the memory (MB) and training energy (mJ) savings per epoch (assuming batch size = 1). The baseline is trained where all weights and BN parameters are tunable. We obtain $\sim95 \%$ and $\sim52 \%$ memory-savings for VGG16/CIFAR10 and ResNet-18/TinyImagenet, respectively. Note, we assume 8-bit precision for convolution weights, activations and their gradients. We achieve $\sim8 \%$  reduction in training-energy for both models, calculated based on the  data in \cite{marinella2018multiscale}.

\section{Conclusion}
\vspace{-1mm}
This work proposes a nearly \textit{training-less} approach for DNN models deployed on memristive crossbars that only optimizes the batchnorm layers in a crossbar-aware manner to preserve the performance of the models in presence of various non-idealities (stochastic read, temporal drift and resistive interconnect parasitics). On hardware, our approach is shown to be highly memory-efficient \& efficient in terms of reducing training energy, as the requirement for hardware-aware fine-tuning of model weights is eliminated.
\vspace{-2mm}


\small
\begin{acks}
\vspace{-1mm}
This work was supported in part by CoCoSys, a JUMP2.0 center sponsored by DARPA and SRC, Google Research Scholar Award, the NSF CAREER Award, TII (Abu Dhabi), the DARPA AI Exploration (AIE) program, and the DoE MMICC center SEA-CROGS (Award \#DE-SC0023198).
\end{acks}
\normalsize
\vspace{-3mm}
\bibliographystyle{ACM-Reference-Format}
\bibliography{sample-base}

\end{document}